\documentclass{amsart}
\usepackage{framed,multirow}

\usepackage{amssymb}
\usepackage{latexsym}
\usepackage{amsmath,amssymb,dsfont,mathtools,enumitem}
\usepackage{url}
\usepackage{xcolor}
\definecolor{newcolor}{rgb}{.8,.349,.1}

\definecolor{LightCyan}{rgb}{0.8,1,1}
\definecolor{LightOrange}{rgb}{1,0.8,0.2}

\DeclareMathOperator{\hl}{hl}
\DeclareMathOperator{\hr}{hr}
\DeclareMathOperator{\vu}{vu}
\DeclareMathOperator{\vl}{vl}
\DeclareMathOperator{\ud}{ud}
\DeclareMathOperator{\ld}{ld}
\DeclareMathOperator{\uad}{uad}
\DeclareMathOperator{\lad}{lad}

\DeclareMathOperator{\oiud}{oiud}
\DeclareMathOperator{\oild}{oild}
\DeclareMathOperator{\oiuad}{oiuad}
\DeclareMathOperator{\oilad}{oilad}
\DeclareMathOperator{\ioud}{ioud}
\DeclareMathOperator{\iold}{iold}
\DeclareMathOperator{\iouad}{iouad}
\DeclareMathOperator{\iolad}{iolad}

\begin{document}

\title[Handwritten character recognition]{Handwritten character recognition using some (anti)-diagonal structural features}

\author[J.M. Casas]{Jos\'e Manuel Casas}
\author[N. Inassaridze]{Nick Inassaridze}
\author[M. Ladra]{Manuel Ladra}
\author[S. Ladra]{Susana Ladra}

\address[1]{Departamento de Matem\'atica Aplicada I, Universidad de Vigo, EUIT Forestal, 36005 Pontevedra, Spain}
\address[2]{A.~Razmadze Mathematical Institute of Tbilisi State University, University Street~2, Tbilisi 0143, Georgia\\ Tbilisi Centre for Mathematical Sciences, Chavchavadze Ave.~75, 3/35, Tbilisi 0168, Georgia}
\address[3]{Departamento de Matem\'aticas, Instituto de Matem\'aticas, Universidade de Santiago de Compostela, 15782 Santiago de Compostela, Spain}
\address[4]{Departamento de Computaci\'on, Universidad de A Coru\~na, Campus de Elvi\~na, 15071 A Coru\~na, Spain}

\begin{abstract}
In this paper, we present a methodology for off-line handwritten character recognition. The proposed methodology relies on a new feature extraction technique based on structural characteristics, histograms and profiles. As novelty, we propose the extraction of new eight histograms and four profiles from the $32\times 32$ matrices that represent the characters, creating 256-dimension feature vectors. These feature vectors are then employed in a classification step that uses a $k$-means algorithm. We performed experiments using the NIST database to evaluate our proposal. Namely, the recognition system was trained using 1000 samples and 64 classes for each symbol and was tested on 500 samples for each symbol.
The accuracy obtained varied from 81.74\% to 93.75\%, showing better results than other methods from the state of the art also based on structural characteristics.
\end{abstract}

\maketitle


\section{Introduction}
Character recognition, popularly referred as optical character recognition (OCR), has been one of the interesting, fascinating and challenging fields of research in pattern recognition, artificial intelligence and machine vision in the last years \cite{MoSuYa, ImOtOc}. It has multiple applications in the real life, such as verifying signatures, recognizing bank check account numbers and amounts, or automating the mail sorting process in postal services; thus, much research has been focused on designing accurate handwritten character recognition systems \cite{BaHaBu, Ho, NaLeSu}. As both industry and academy have paid attention to this attractive field, there have been numerous previous attempts for recognizing handwritten characters, such as those methods for English handwritten characters included in \cite{KuTa, MoSuYa, NaLeSu, ShaGhShaTh}. One can find information about recent trends and tools in OCR in \cite{ShaGhShaTh}.

The generic handwritten recognition process includes several phases:  preprocessing the handwritten text, segmenting the writing into isolated characters, extracting feature vectors from the individual characters, and finally classifying each character using the features previously extracted, such that it can be assigned to the most likely letter \cite{Ho}.

In this work, we will focus on the feature extraction step, as it is vital for obtaining good results in the recognition process in terms of accuracy. Its main goal is to extract and select a collection of features that maximizes the recognition rate using the least amount of elements as possible \cite{ArYa}.
Moreover, the feature extraction method should be robust, that is, it should obtain similar feature sets from a collection of instances of the same symbol. This property makes the subsequent classification step less difficult \cite{FiGeKe}.



According to Govindan and Shivaprasad \cite{GoShr}, we can classify features in different categories: global transformation and series expansion, statistical features, and geometrical and topological features.
This last approach allows encoding some knowledge
about the structure of the object and the components
that make up that object \cite{ArYa}.
In addition, it is closer to the human way of recognition \cite{BuSa}.
The structural features consider different properties of the characters, such as extreme points, maxima and minima, reference lines, ascenders, descenders, cusps above and below a threshold, isolated dots, cross points, branch points, direction of a strokes, inflection between two points, etc \cite{ArYa}. Some of these structural features were already proposed by cognitive psychologists \cite{LiNo}, when studying the visual and cognitive
mechanisms involved in visual object recognition.

There are many works in OCR using structural feature extraction models.
Rocha and Pavlidis \cite{RoPa} proposed a method for the recognition of multi-font printed characters giving special emphasis to structural features. The structural description of the shape for each character considered convex arcs and strokes, singular points and their spatial interrelations.
Kahan et al.~\cite{KaPaBa} also developed a structural feature set for recognition of printed text. They included different information for a character, such as the location and number of their holes, the concavities in their skeletal structure, characteristics of their bounding boxes, among others.
Kuroda, Harada, and Hagiwara implemented
a recognition system for the on-line identification of handwritten Chinese characters based on structural patterns \cite{KuHaHa}.
Lee and Gomes \cite{LeGo} proposed a method for recognizing numeral characters, also based on structural features. Their technique considered topological characteristics like the number of cavities, the crossing sequences, the intersection with the principal and secondary axes, and the distribution of pixels.
Chan and Yeung \cite{ChYe} proposed a structural approach for the analysis of handwritten mathematical expressions. This problem is even more complicated than recognizing individual characters or symbols, as the components of the mathematical expression are normally arranged as a complex two-dimensional
structure, and have different sizes. Amin \cite{Am} focused on printed Arabic text, which obtains lower recognition rates than those of disconnected
characters such as printed English. He used seven types of global structural features such as number of sub words, number of peaks of each word, number of loops of each peak, number and position of complimentary characters, or the height and width of each peak.
Kavallieratou, Fakotakis and Kokkinakis \cite{KaFaKo_2} proposed an integrated analysis system for unconstrained handwriting. The last module of this system includes a handwritten character recognition technique that uses a structural approach \cite{KaFaKo}. More concretely, it extracts a 280-dimensional feature vector for each character, consisting of the horizontal, vertical and radial histograms and the out-in and in-out radial profiles, and uses the $k$-means algorithm for the classification. Yang, Lijia and Chen \cite{YaLiChe} proposed the combination of structural and statistical features, in addition to BP networks for the classification step, to solve interferences of external noise.\\

In this paper, we propose a new algorithm for isolated English handwritten character recognition based on some structural features, using eight new histograms and four new profiles. Thus, we extract a $256$-dimensional integral vector for each character and then employ the $k$-means clustering algorithm for the classification step. We compare our results to those given in \cite{KaFaKo}, as the methods for feature extraction and classification are the most similar to ours. Our illustrative tables show that we reduce the dimension of the feature vectors and improve the accuracy of recognition.

The rest of the paper is organized as follows. Section~\ref{S:algo} describes the proposed technique in detail. Section~\ref{S:exp} includes the experimental evaluation, and finally Section~\ref{S:concl}  contains the conclusions and future work.

\section{Proposed algorithm}\label{S:algo}

\subsection{Preliminaries}
A handwritten character recognition system usually requires a preprocessing phase before the feature extraction and classification steps \cite{KuTa}. The main goal of this preprocessing phase is to obtain isolated characters and represent them conveniently for the following steps. In most cases, this includes a segmentation stage and a binarization stage to get the isolated characters in the form of $m \times n$ binary matrices.
These matrices are then generally normalized by reducing the size and removing the redundant information from the image without loosing any important information.

Then, the feature extraction is applied over these matrices. This step can be considered as the heart of the system, as the feature selection is usually the most important factor to achieve high accuracies in the recognition process. After the normalization of the character images, the objective of the feature extraction is to represent the isolated characters as unique feature vectors. The key is to maximize the recognition rate using as few features as possible.

Finally, the classification stage is the main decision-making stage of the system, and it uses the feature vectors to identify the text segment according to preset rules.
In this stage, the basic task is to design a decision rule that is easy to compute and that maximizes the certainty of the misclassification relative to the power of the feature extraction scheme employed.

\subsection{Preprocessing}
For the experimental evaluation, we plan to use NIST database \cite{NIST}, which contains $128 \times 128$ BMP files for isolated handwritten English characters. Thus, before extracting the features, the preprocessing step must binarize and then normalize each original image data file to obtain a $32 \times 32$ matrix with entries in $\{0,1\}$, such that $0$s stand for white pixels while $1$s for black pixels.

\subsection{Features extraction}
As we have already mentioned, in this paper we focus on structural characteristics for feature extraction.
Instead of the well-known horizontal and vertical histograms, we introduce new horizontal left and right histograms and vertical upper and lower histograms. We also employ new orthodiagonal and orthoantidiagonal histograms and profiles. All these features are used for the first time in the optical character recognition research. We will study if these new features improve the accuracy of the handwritten character recognition algorithm in comparison with \cite{KaFaKo}.\\

Now, we give the formal definition of these features. We need a map
\[
f \colon [32] \times [32] \longrightarrow \{0, 1\}
\]
defined as follows: $f(l,m)$ is the value of the element in the $l$-th row and $m$-th column of the character matrix, and $[32] = \{1, \dots , 32\}$.

The horizontal left and right histograms, $H_{\hl}$ and $H_{\hr}$, of the character matrix are the number of black pixels in the even rows of the left half of the matrix and the odd rows of the right half of the matrix, respectively (i.e. 32 features):
\[
H_{\hl}(n)= \displaystyle  \sum^{16}_{m=1}  f(2n,m) \quad \text{for all} \quad 1\leq n \leq 16,
\]
and
\[
H_{\hr}(n)= \displaystyle  \sum^{32}_{m=16}  f(2n-1,m) \quad \text{for all} \quad 1\leq n \leq 16.
\]

The vertical upper and lower histograms, $H_{\vu}$ and $H_{\vl}$,  of the character matrix are the number of black pixels in the even columns of the upper half of the matrix and the odd columns of the lower half of the matrix, respectively (i.e. 32 features):
\[
H_{\vu}(n)= \displaystyle \sum^{16}_{m=1}    f(m,2n) \quad \text{for all} \quad 1\leq n \leq 16,
\]
and
\[
H_{\vl}(n)= \displaystyle  \sum^{32}_{m=16}   f(m,2n-1) \quad \text{for all} \quad 1\leq n \leq 16.
\]

Besides above-given histograms, we introduce several other new histograms. We start with the upper and lower diagonal histograms, $H_{\ud}$ and $H_{\ld}$, given by the number of black pixels according to the odd and even orthogonal lines to the diagonal of the character matrix in the upper and lower triangles, respectively (i.e. 32 features in total):
\[
H_{\ud}(n)= \displaystyle  \sum_{\substack{k\geq 0\\ 2n-1-k\geq 1 \\ 2n-1+k\leq 32}}  f(2n-1-k,2n-1+k) \quad \text{for all} \quad 1\leq n \leq 16,
\]
and
\[
H_{\ld}(n)= \displaystyle  \sum_{\substack{k\geq 0\\ 2n-k\geq 1 \\ 2n+k\leq 32}}   f(2n+k,2n-k) \quad \text{for all} \quad 1\leq n \leq 16.
\]

Symmetrically, the upper and lower antidiagonal histograms, $H_{\uad}$ and $H_{\lad}$, are defined as the number of black pixels according to the even and odd orthogonal lines to the antidiagonal of character matrix in upper and lower triangles, respectively (i.e. again 32 features in total):
\[
H_{\uad}(n)= \displaystyle  \sum_{\substack{k\geq 0\\ 2n-k\geq 1 \\ 33-2n-k\geq 1}}  f(2n-k,33-2n-k) \quad \text{for all} \quad 1\leq n \leq 16,
\]
and
\[
H_{\lad}(n)= \displaystyle  \sum_{\substack{k\geq 0\\ 2n+k-1\leq 32 \\ 34-2n+k\leq 32}}  f(2n-1+k,34-2n+k) \quad \text{for all} \quad 1\leq n \leq 16.
\]

Additionally, we introduce the out-in and in-out diagonal and antidiagonal profiles for each normalised character. Namely, \emph{out-in upper diagonal profile} $P_{\oiud}$ and \emph{out-in lower diagonal profile} $P_{\oild}$  are defined at the index $1\leq n \leq 16$ as the position of the first black pixel found in the $(2n-1)$-th orthogonal line to the diagonal of the character matrix, starting from the periphery in the upper triangle going down; and the $2n$-th orthogonal line to the diagonal of the character matrix, starting in lower triangle going up, respectively (i.e. 32 features in total):
\[
P_{\oiud}(n) = \left\lbrace    I \ \left|
  \begin{array}{c}
  \displaystyle  \sum_{\substack{k\geq I+1\\ 2n-1-k\geq 1 \\ 2n-1+k\leq 32}} f(2n-1-k,2n-1+k) = 0 \\
\qquad \qquad  f(2n-1-I, 2n-1+I)=1 \\
 \end{array} \right. \right.
\]
for all $1\leq n \leq 16$, and

\[
P_{\oild}(n) = \left\lbrace    J \ \left|
  \begin{array}{c}
  \displaystyle  \sum_{\substack{k\geq J+1\\ 2n-k\geq 1 \\ 2n+k\leq 32}} f(2n+k,2n-k) = 0 \\
\qquad \qquad  f(2n+J, 2n-J)=1 \\
 \end{array} \right. \right.
\]
for all $1\leq n \leq 16$.

Symmetrically, we introduce the \emph{out-in upper antidiagonal profile} $P_{\oiuad}$ and the \emph{out-in lower antidiagonal profile} $P_{\oilad}$, which  are defined at the index $1\leq n \leq 16$ as the position of the first black pixel found in the $2n$-th orthogonal line to the antidiagonal of the character matrix, starting from the periphery in upper triangle going down; and the $(2n-1)$-th orthogonal line to the antidiagonal of the character matrix, starting in lower triangle going up, respectively (i.e. 32 features in total):
\[
P_{\oiuad}(n) = \left\lbrace    I \ \left|
  \begin{array}{c}
   \displaystyle  \sum_{\substack{k\geq I+1\\ 2n-k\geq 1 \\ 33-2n-k\geq 1}} f(2n-k,33-2n-k) = 0 \\
\qquad \qquad f(2n-I, 33-2n-I)=1 \\
 \end{array} \right. \right.
\]
for all $1\leq n \leq 16$, and

\[
P_{\oilad}(n) = \left\lbrace    J \ \left|
  \begin{array}{c}
   \displaystyle  \sum_{\substack{k\geq J+1\\ 2n-1+k\leq 32 \\ 34-2n+k\leq 32}} f(2n-1+k,34-2n+k) = 0 \\
\qquad \qquad \ f(2n-1+J, 34-2n+J)=1 \\
 \end{array} \right. \right.
\]
for all $1\leq n \leq 16$.

\;

Moreover, the \emph{in-out upper diagonal profile} $P_{\ioud}$ and the \emph{in-out lower diagonal profile} $P_{\iold}$ are defined at the index $1\leq n \leq 16$ as the position of the first black pixel found in the $(2n-1)$-th and in the $2n$-th orthogonal lines to the diagonal of character matrix starting from the diagonal going to the periphery in the upper and lower triangles, respectively (i.e. 32 features in total):
\[
P_{\ioud}(n) = \left\lbrace    I \ \left|
  \begin{array}{c}
  \displaystyle  \sum^{I-1}_{\substack{k\geq 0\\ 2n-1-k\geq 1 \\ 2n-1+k\leq 32}} f(2n-1-k,2n-1+k) = 0 \\
\qquad \qquad  f(2n-1-I, 2n-1+I)=1 \\
\end{array} \right. \right.
\]
for all $1\leq n \leq 16$, and

\[
P_{\iold}(n) = \left\lbrace    J \ \left|
  \begin{array}{c}
  \displaystyle  \sum^{J-1}_{\substack{k\geq 0\\ 2n-k \geq 1 \\ 2n+k\leq 32}} f(2n+k,2n-k) = 0 \\
\qquad \quad f(2n+J, 2n-J)=1 \\
 \end{array} \right. \right.
\]
for all $1\leq n \leq 16$.

Symmetrically, we introduce the \emph{in-out upper antidiagonal profile} $P_{\iouad}$ and the \emph{in-out lower antidiagonal profile} $P_{\iolad}$, which are defined at the index $1\leq n \leq 16$ as the position of the first black pixel found in the $2n$-th and in the $(2n-1)$-th orthogonal lines to the antidiagonal of the character matrix, starting from the antidiagonal going to the periphery in upper and lower triangles, respectively (i.e. 32 features in total):

\begin{table*}[!th]
 \centering
  \caption{\label{Ta:TrSet}Training set from NIST database}
\begin{tabular}{|l|c|c|}
	\hline
	NIST database  & Partition & Handwriting Sample Forms \\
	\hline
	Digits  & $HSF_0$ & $F0000 \ldots F0099$ \\
	\hline
	Uppercase characters & $HSF_0$, $HSF_1$ & $F0000 \ldots F0999$ \\
	\hline
	Lowercase characters & $HSF_0$, $HSF_1$ &  $F0000 \ldots F0999$\\
	\hline
\end{tabular}
\end{table*}

\;

\begin{table*}[!th]
 \centering
   \caption{\label{Ta:TstSet}Test set from NIST database}
  \begin{tabular}{|c|c|c|}
	\hline
	NIST database  & Partition & Handwriting Sample Forms \\
	\hline
	Digits  & $HSF_0$ & $F0100 \ldots F0149$ \\
	\hline
	Uppercase characters & $HSF_3$ & $F1000 \ldots F1499$ \\
	\hline
	Lowercase characters & $HSF_3$ &  $F1000 \ldots F1499$\\
	\hline
\end{tabular}
\end{table*}

\begin{table*}[!th]
  \centering
  \caption{\label{Ta:Ka}Results of Algorithm~\cite{KaFaKo}}
\begin{tabular}{|c|c|c|c|}
	\hline
	 & 1\textsuperscript{st} Choice & 2\textsuperscript{nd} Choice & 3\textsuperscript{rd}  Choice \\
	\hline
	Digits & 92.48\% & 96.02\% & 97.60\% \\
	\hline
 	Uppercase characters & 87.08\% & 92.95\% & 95.26\% \\
	\hline
	Lowercase characters & 79.71\% & 88.62\% & 92.12\% \\
	\hline
\end{tabular}
\end{table*}

\;

\begin{table*}[!th]
  \centering
    \caption{\label{Ta:Niko}Results of our Algorithm}
  \begin{tabular}{|c|c|c|c|}
	\hline
	 & 1\textsuperscript{st} Choice & 2\textsuperscript{nd} Choice & 3\textsuperscript{rd}  Choice \\
	\hline
	Digits & 93.75\% & 97.02\% & 97.90\% \\
	\hline
 	Uppercase characters & 88.58\% & 94.09\% & 95.79\% \\
	\hline
	Lowercase characters & 81.74\% & 90.13\% & 92.89\% \\
	\hline
\end{tabular}
\end{table*}

\[
P_{\iouad}(n) = \left\lbrace    I \ \left|
  \begin{array}{c}
\displaystyle  \sum^{I-1}_{\substack{k\geq 0\\ 2n-k \geq 1 \\ 33-2n-k \geq 1}} f(2n-k,33-2n-k)= 0 \\
\qquad  \qquad  f(2n-I, 33-2n-I)=1 \\
 \end{array} \right. \right.
\]
for all $1\leq n \leq 16$, and

\[
P_{\iolad}(n) = \left\lbrace    J \ \left|
  \begin{array}{c}
\displaystyle  \sum^{J-1}_{\substack{k\geq 0\\ 2n-1+k\leq 32 \\ 34-2n+k\leq 32 }} f(2n-1+k,34-2n+k) = 0 \\
\qquad  \qquad \  f(2n-1+J, 34-2n+J)=1 \\
 \end{array} \right. \right.
\]
for all $1\leq n \leq 16$.

\subsection{Classification}
In the previous step, a $256$-dimensional feature vector have been extracted from each isolated handwritten character image.
These feature vectors are then used in the classification step, where we use the $k$-means clustering algorithm to train and create a classification model.

\section{Experimental evaluation} \label{S:exp}
We run experiments using the NIST database of handwritten English characters \cite{NIST}. The experiments were held separately for each one of the following categories: digits, uppercase characters and lowercase characters.

In more detail, using programs written in Python, our recognition algorithm was trained and tested in comparison with the algorithm given in \cite{KaFaKo} on about 1000 samples and 64 classes and on 500 samples for each isolated handwritten character from NIST database, respectively.  Namely, Table~\ref{Ta:TrSet} and Table ~\ref{Ta:TstSet} show the exact input data for our experiments.

Thus, the training and the test set of our experiments were completely disjoint, which means that the writers used in testing were completely different from the ones used for training.

We show the accuracy rate obtained by \cite{KaFaKo} for each character category in Table~\ref{Ta:Ka}, whereas Table~\ref{Ta:Niko} shows the accuracy rates obtained by our method. Analogously to \cite{KaFaKo}, we also show the recognition accuracy rates when the second and third choices are taken into account, as in a real system one could use lexicons to improve the output of the proposed character recognizer. These results show that our technique outperforms the algorithm in \cite{KaFaKo} for all categories, in some cases by a margin of more than 2 percentage points.

\section{Conclusion}\label{S:concl}

In this paper, we present a technique for English handwritten character recognition based on the extraction of new structural features. More concretely, we introduce eight new histograms and four new profiles, which have been proven to successfully represent the handwritten characters.

We have tested our approach using the NIST database and obtained recognition accuracies varying
from 81.74\% to 93.75\%, depending on the difficulty of the character category. These results outperform previous attempts of using just structural features, in addition to being fast and simple to compute.

The results are promising and usable is some sort of applications. In the nearest future they will be implemented in a mobile (iOS) application. We also plan to apply the technique to characters from other languages, e.g. Georgian characters. Moreover, due to the nature of our method, it is possible to reduce the current number of features, 256 in this paper, according to the needs of the application where the technique will be used.

\section*{Acknowledgments}
This work was supported by the Agencia Estatal de Investigaci\'on, Spain (European ERDF support included, UE) [grant numbers MTM2016-79661-P, TIN2016-77158-C4-3-R]; and by the Conseller\'ia de Cultura, Educaci\'on e Ordenaci\'on
Universitaria and the European Regional Development Fund (ERDF) [grant number ED431G/01].


\bibliographystyle{model2-names}

\begin{thebibliography}{99}

\bibitem{Am} A. Amin, ``Recognition of printed Arabic text based on global features and decision tree learning techniques'', Pattern Recognition, Vol. 33, pp. 1309--323, 2000.

\bibitem{ArYa}  N. Arica and F. Yarman-Vural, ``An Overview of Character Recognition Focused on  Off-line Handwriting'', IEEE Transactions on Systems, Man, and Cybernetics, Part C: Applications and Reviews, Vol. 31, No. 2, pp. 216--233, 2001.

\bibitem{BaHaBu} C. Bahlmann, B. Haasdonk, H. Burkhardt, ``Online Handwriting Recognition with Support Vector Machine – A Kernel Approach'',
In proceeding of the 8th Int. Workshop in Handwriting Recognition (IWHFR), pp. 49--54, 2002.

\bibitem{Ho} H.S.M. Beigi, ``An Overview of Handwriting Recognition'', Proceedings of the 1st Annual Conference on Technological Advancements in Developing Countries, Columbia University, New York, pp. 30--46, 1993.

\bibitem{BuSa} H. Bunke, A. Sanfeliu, ``Syntactic and structural Pattern Recognition'', Theory and Applications, World Scientific, Singapore, 1990.

\bibitem{ChYe}  K.-F. Chan, D.-Y. Yeung, ``An efficient syntactic approach to structural analysis of on-line handwritten mathematical expressions'', Pattern Recognition, Vol. 33, pp. 375--384, 2000.

\bibitem{FiGeKe} J.A. Fitzgerald, F. Geiselbrechtinger, T. Kechadi, ``Application of fuzzy logic to online recognition of handwritten 	symbols'', in: Proceedings of the Ninth
International Workshop on Frontiers in Handwritten 	Recognition, pp. 395-–400, 2004.

\bibitem{GoShr} V.K. Govindan, A.P. Shivaprasad,``Character Recognition-A review'', Pattern Recognition, vol. 23, no. 7, pp. 671--683, 1990.

\bibitem{ImOtOc} S. Impedovo, L. Ottaviano, S. Occhinegro, ``Optical character recognition'', International Journal Pattern Recognition and Artificial Intelligence, vol. 5(1-2), pp. 1--24, 1991.


\bibitem{KuTa}   A.K. Jain, T. Taxt, ``Feature extraction Methods for Character Recognition- A Survey'', Pattern Recognition, vol. 29, no. 4, pp. 641--662, 1996.

\bibitem{KaPaBa} S. Kahan, T. Pavlidis, H.S. Baird, ``On the recognition of printed characters of any font and size'',  IEEE Transactions on Pattern Analysis and Machine Intelligence, vol. PAMI-9, no. 2, pp. 274--288, 1987.

\bibitem{KaFaKo} E. Kavallieratou, N. Fakotakis, G. Kokkinakis, ``Handwritten Character Recognition based on Structural Characteristics'', In
 Proceedings of the 16th International Conference of Pattern Recognition ICPR 2002,  IEEE, vol. III, pp. 139--142, Quebec-Canada, 2002.

\bibitem{KaFaKo_2} E. Kavallieratou, N. Fakotakis, G. Kokkinakis, ``An unconstrained handwriting recognition system'', International Journal on Document Analysis and Recognition, vol. 4, no. 4, pp. 226--242, 2002.

\bibitem{KuHaHa}
K. Kuroda, K. Harada, M. Hagiwara, ``Large Scale On-Line Handwritten Chinese Character Recognition Using Improved Syntactic Pattern Recognition'', Proceedings of the IEEE International Conference on Systems, Man, and Cybernetics, vol. 5, pp. 4530 -- 4535, 1997.

\bibitem{LeGo} L.L. Lee, N.R. Gomes, ``Disconnected handwritten numeral image recognition'',
in the Proceedings of 4-th ICDAR, pp. 467--470, 1997.

\bibitem{LiNo} P.H. Lindsay, D.A.  Norman, ``Human information
processing: An introduction to psychology'',
New York: Academic Press, 1977.


\bibitem{MoSuYa} S. Mori, C.Y. Suen, K. Yamamoto, ``Historical review of OCR research and development'', Proc. of IEEE, vol. 80, pp. 1029--1058, 1992.

\bibitem{NaLeSu} C. Nadal,  R. Legault,  C.Y Suen, ``Complementary Algorithms for Recognition of totally Unconstrained Handwritten Numerals'', in Proc. 10th Int. Conf. Pattern Recognition,  vol. 1, pp. 434--449, 1990.


\bibitem{ShaGhShaTh} Om Prakash Sharma, M.K. Ghose, Krishna Bikram Shah, Benoy Kumar Thakur, ``Recent Trends and Tools for Feature Extraction in OCR Technology'', Proceeding of the International Journal of Soft Computing and Engineering, vol. 2, no. 6, pp. 220--223, 2013.

\bibitem{RoPa} J. Rocha, T. Pavlidis, ``A shape analysis model with applications to a character recognition system'', IEEE Transactions on PAMI, Vol. 16(4), pp. 393--404, 1994.

\bibitem{NIST} R. Wilkinson, J. Geist, S. Janet, P. Grother, C. Burges, R. Creecy, B. Hammond, J. Hull, N. Larsen, T. Vogl,  C. Wilson, The first census optical character recognition systems conf. \#NISTIR 4912. The U.S Bureau of Census and the National Institute of Standards and Technology. Gaithersburg, MD, 1992.

\bibitem{YaLiChe} Yang Yang, Xu Lijia, Cheng Chen, ``English character recognition based on feature combination'', Procedia Engineering, Vol. 24, pp. 159-–164, 2011.
\end{thebibliography}

\end{document}